\documentclass[10pt, a4paper]{article}

\usepackage{lrec-coling2024} % this is the new style
\usepackage{physics}
\usepackage{mathtools,nccmath}
\usepackage{graphicx}
\usepackage{booktabs}
\usepackage{amsfonts,amssymb}
\usepackage{bm}
\usepackage{threeparttable}
\usepackage{multirow}
\usepackage{amsmath}
\usepackage{arydshln}
\usepackage{subfigure}
\usepackage{enumitem}
\usepackage{times}
\usepackage{soul}
\usepackage{url}
\usepackage{amsthm}
\usepackage{booktabs}
\usepackage{algorithm}
\usepackage{algorithmic}
\usepackage{xcolor}
\usepackage{caption}
\usepackage{CJKutf8}
\usepackage{hyperref}
\usepackage{colortbl} % For cell coloring
\usepackage{xcolor}
\usepackage{authblk}
\definecolor{Gray}{gray}{1} % Define the gray color

\title{PopALM: Popularity-Aligned Language Models for Social Media\\Trendy Response Prediction }

\name{Erxin Yu\textsuperscript{1}, Jing Li\textsuperscript{1,2}$^{\ast}$\thanks{*Corresponding author}, Chunpu Xu\textsuperscript{1}} 

\address{\textsuperscript{1}Department of Computing, The Hong Kong Polytechnic University\\
        \textsuperscript{2}Research Centre for Data Science \& Artificial Intelligence\\
         \{erxin.yu, chun-pu.xu\}@connect.polyu.hk, 
         jing-amelia.li@polyu.edu.hk\\}

\abstract{

Social media platforms are daily exhibiting millions of events.
To preliminarily predict the mainstream public reaction to these events, we study \textit{trendy response prediction} to automatically generate top-liked user replies to social media events. 
While previous works focus on generating responses without factoring in popularity, we propose \textbf{Pop}ularity-\textbf{A}ligned \textbf{L}anguage \textbf{M}odels (\textbf{PopALM}) to distinguish responses liked by a larger audience through reinforcement learning.
Recognizing the noisy labels from user ``likes'', we tailor-make curriculum learning in proximal policy optimization (PPO) to help models capture the essential samples for easy-to-hard training.
In experiments, we build a large-scale Weibo dataset for trendy response prediction, and its results show that PopALM can help boost the performance of advanced language models.

 \\ \newline  }

\begin{document}

\maketitleabstract

\section{Introduction}
Social media is a popular channel for users to voice opinions and share information, making it an asset for studying real-world events on diverse topics and public views of them.
It is a valuable resource for analyzing and predicting events' mainstream social responses, benefiting various applications, e.g., early event analysis, public response simulation, and comment generation ~\citep{CommnetGenerationAAAI21, CommnetGenerationTOMM2022}.
However, the vast volumes of daily-created events are beyond humans' ability to track each.
Therefore, we study trendy response prediction to automate the generation of top-liked user responses, which can helpfully train language models to predict the mainstream public reaction before an event happens or in its early stages.
Here, response popularity is characterized by how many users ``like'' it, where \textit{like} is a social media behavior showing an audience's agreement to a response \citep{gao-etal-2020-dialogue}.

Despite the breakthrough progress in automatic response generation thanks to the advances in large language models (LLMs) \citep{instructGPTNIPS2022}, most previous work focuses on generic human responses without considering the popularity factors in the social contexts.
However, compared to generic responses, popular responses are much more closely linked to the events' trajectory \citep{ding-etal-2020-hashtags} and better reflect the mainstream voices of the public \citep{kano-etal-2018-harnessing}. 

\begin{figure}
    \centering  \includegraphics[width=.5\textwidth]{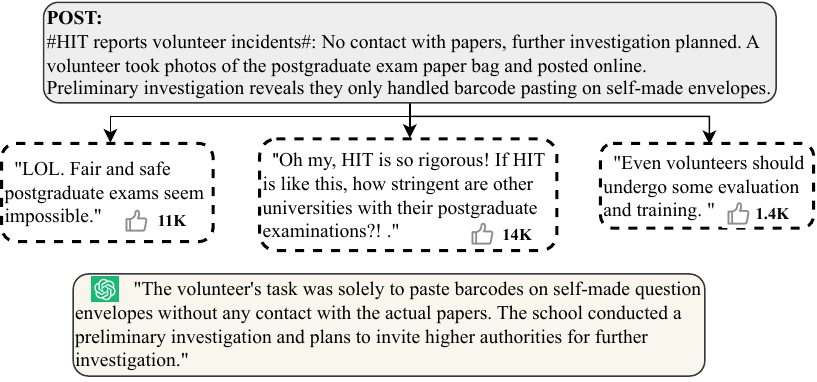}
    \caption{\label{fig:intro-case}
    A Weibo post about ``\textit{Volunteer Leaked Exam Questions}'', followed by its top-3 trendy responses with like numbers. The bottom presents a response sample generated by ChatGPT. 
    % Here we show the English translation and put the original content in Chinese in Appendix \ref{appendix-Chinese}.
    }
    %\vspace{-1em}
    
\end{figure}

To illustrate this point, Figure \ref{fig:intro-case} shows a societal event example about ``\textit{Volunteer Leaked Exam Questions}'' with its description post from Weibo (a Chinese social media platform) and the top-3 trendy responses by audiences' like numbers; we also display ChatGPT's prediction about the possible trendy response for comparison. 
As can be seen, the real trendy responses can better reflect people's opinions and emotions, e.g., surprise at the leakage of exam papers and doubts about examination fairness. 
In contrast to these specific points, the output of ChatGPT focuses on a macro level, hence inferior in reflecting essential and concrete public viewpoints.

Given these concerns, we propose  Popularity-Aligned Language Models (PopALM) to train language models with popularity via reinforcement learning.
To the best of our knowledge, \emph{PopALM exhibits the first effort to align language generation with social media popularity measure.}
We adopt like numbers to train the reward function and employ a PPO method to optimize the training process.
However, like numbers, although as easy-to-access popularity indicators, are noisy user-generated labels, which may be affected by many factors beyond text, such as posting time, authors, etc.
These noisy labels may thus exhibit implicit relations to the text features, substantially challenging the training of reward functions.

To address this challenge, PopALM engages curriculum learning \cite{CurriculumLearningICML2009} into PPO to filter out the noisy training samples and differentiate the samples' learning difficulty for optimizing the learning pace from easy to hard.
First, the reward function leverages task-specific supervision to align with trendy response prediction.
Then, we rank the samples based on the reward prediction confidence to remove noisy samples, i.e., samples with low confidence.
Lastly, we employ the self-paced learning strategy for the remaining samples to progressively learn from easy to hard samples, thus improving the overall learning efficiency.

As a pilot study on trendy response prediction, we should benchmark the task with the first dataset. 
To that end, we collect around 30K daily-trending events from Weibo, each with the most popular post as its description.
% \footnote{\url{weibo.com}}
To explore trendy responses for each post, we also gather its user replies associated with the like numbers for popularity learning.
\begin{figure*}
    \centering
    \includegraphics[width=.9\textwidth]{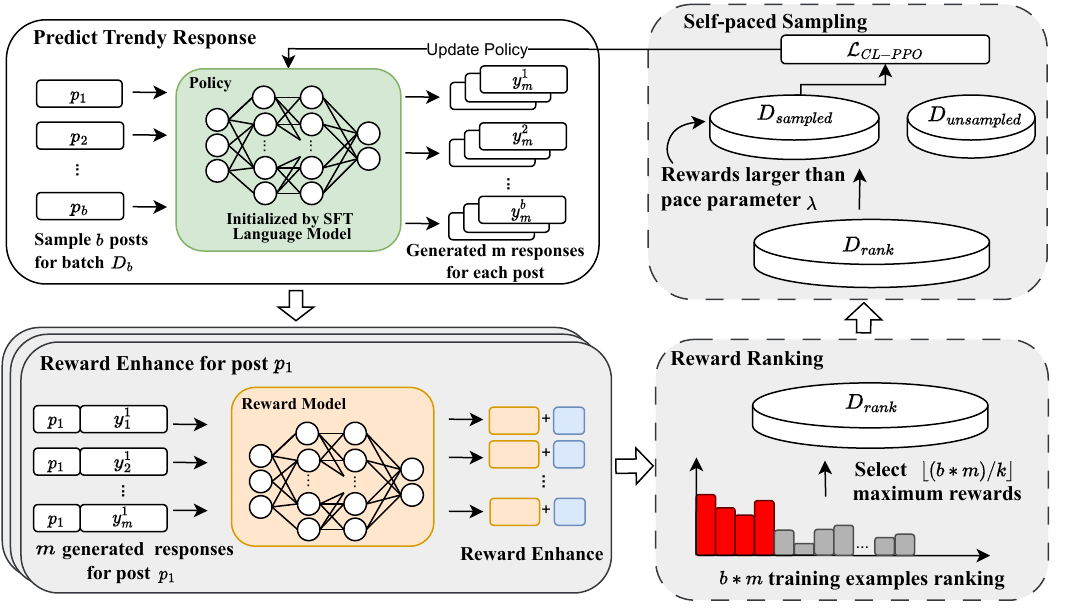}
    %\vspace{-0.5em}
    \caption{ \label{fig: framework}The workflow of PopALM is based on curriculum learning enhanced PPO, which exploits three novel strategies to leverage noisy user-like labels as popularity indicators.
    These strategies are Reward Enhancement (left bottom; for task-specific supervision), reward ranking (right bottom; for filtering noisy training samples), and self-paced reward sampling (right top; for training from easy to hard). }
    \vspace{-0.5em}
   
\end{figure*}
The main comparison results in experiments demonstrate that PopALM helps advanced language generation models improve trendy response prediction quality in both automatic and human evaluation.
Then, ablation studies indicate the positive contributions of curriculum learning strategies to PopALM's overall effectiveness.
Next, quantitative analysis shows PopALM's superiority with varying training models and data scales.
Finally, we demonstrate the enhancing effect of the generated responses on other tasks.
% and an error analysis.

In summary, our contributions are three-fold:

$\bullet$ We present the first study on trendy response prediction for social media events and build the first large-scale benchmark for this task.

$\bullet$ We propose PopALM, a novel popularity-aligned language model, with a tailor-made curriculum learning for PPO to effectively learn from noisy user-like labels in social media contexts.

$\bullet$ We extensively experiment on PopALM and find that it works well in different LLMs and fine-tuning methods to learn popularity and help predict trendy responses.

\section{Related Work}\label{sec:related-work}

\paragraph{Response Generation.}
% Response Generation has received more and more attention in the NLP community.
Our task is in line with response generation, an increasingly popular field in NLP.
Its early work applied the RNN-based sequence-to-sequence model and achieved promising results \citep{shang-etal-2015-neural,responseGenAAAI2018,zhang-etal-2019-recosa}. 
In recent years, pre-trained large-scale language models have brought many breakthroughs in natural language generation, e.g., the GPT series \citep{gpt1,gpt2,gpt3}, T5 \citep{T5}, and BART \citep{lewis-etal-2020-bart}.

Building upon these models, numerous methods have been proposed to enhance response generation capabilities.
DialoGPT \citep{zhang-etal-2020-dialogpt} is tailored for response generation using comments sourced from Reddit.
The blender model \citep{roller-etal-2021-recipes} refines the pre-trained model using responses annotated by humans to emphasize desired conversational capabilities, such as engagement, knowledge, empathy, and personality.
PLATO \citep{bao-etal-2020-plato} introduces discrete latent variables to address the inherent one-to-many mapping problem to improve response quality.

In this field, many studies also focused on automatic comment generation in a social media context \citep{autoCommentGen1,qin-etal-2018-automatic}.
%Automatic comment generation is a subfield of response generation that aims to generate generic comments \citep{autoCommentGen1,qin-etal-2018-automatic}. 
\citet{yang-etal-2019-read} allowed better encoding through selected important contextual spans. 
\citet{autoGenTOMM} leveraged the topic model to capture the author's styles for a personalized generation.
However, most of these studies focused on generating generic or individual' comments, paying limited attention to trendy response generation with popularity measures, revealing a gap to address.
%enhance the quality of generated comments.
% Works in this field have not yet considered the comment's popularity. 
%Recent works are still built upon sequence-to-sequence frameworks. \citept{yang-etal-2019-read} focus on selecting important spans from the news article. \citept{autoCommentGenAAAI} leverage the topic model to enhance the quality of generated comments. \citept{autoGenTOMM} 
% consider the user linguistic style modeling for personalized comment generation.

\paragraph{Language Models Alignment.}
Taking recent advances in large language models (LLMs), many studies examined how to align language models with human feedback \citep{summary,learningFromHumanFeedback}.
For example, ChatGPT, a closely related model to InstructGPT \citep{instructGPTNIPS2022}, is specifically trained to follow human instructions and has demonstrated state-of-the-art performance in conversational abilities.
ChatGLM is a bilingual language model that aligns the General Language Model \citep{du-etal-2022-glm} with large-scale human instructions, achieving superior performance in the Chinese response generation.
%In this work, we study the problem of trendy response generation by 
PopALM can be based on various language models and explores trendy responses with popularity alignment, which has not been explored previously.

\paragraph{Popularity Prediction.}
Our work is related to popularity prediction on social media, where users express their preferences by voting, sharing, or bookmarking a post. 
The count of such actions is usually adopted as the popularity indicator. \citet{lamprinidis-etal-2018-predicting} used a multi-task GRU network to predict headline popularity. \citet{kano-etal-2018-harnessing} employed such popularity measure to distantly supervise extractive summarization. 
\citet{gao-etal-2020-dialogue} leveraged social media feedback data (number of replies and upvotes) to build a large-scale dataset to classify user feedback.
However, none of them injects the popularity factors into language generation, which we will extensively explore.
\section{Popularity-Aligned Language Models}

\paragraph{PopALM Overview.} To begin with, we state the problem of trendy response prediction as follows: given post $p$, the model needs to generate trendy responses $Y = \{y_1, y_2,..., y_m\}$, in which $y_i$ is one of the popular responses.
As shown in Figure \ref{fig: framework} (the workflow to build PopALM), following InstructGPT \cite{instructGPTNIPS2022}, our framework consists of three parts: supervised fine-tuning, reward modeling, and reinforcement learning (RL). 
Our RL algorithm is based on PPO, and we further introduce curriculum-learning engaged PPO (CL-PPO) to alleviate the noisy labels challenge in the popularity learning of social media. 
We describe our RL-based backbone framework in Section \ref{ssec:ppo}, followed by our CL-PPO algorithm in Section \ref{ssec:cl-ppo}. 

\subsection{Aligning LMs with Popularity via RL}\label{ssec:ppo}

\paragraph{Supervised Fine-tuning.}
We first fine-tune language models (LMs) to predict trendy responses using supervised learning. 
In this stage, we only consider the one-to-one mapping relation between one post and a trendy response. 
Given one post $p$ and its trendy responses $Y$, we pair $p$ with each response in $Y$, forming our supervised training samples $\{(p, y_1), (p, y_2)..., (p, y_m) \}$. 
Here, the training object for one post is to minimize the following negative log-likelihood (NLL) loss:
   
   \begin{equation} \small
        \mathcal{L}_{SFT} = -E_{(p,y_i)\sim D_{SFT}}\sum_{i=1}^{m}\sum_{t=1}^{T} -\log p(y^{t}_{i}|p,y^{<t}_{i}),
   \end{equation}

\noindent where T is the length of the response, $D_{SFT}$ is the dataset for supervised fine-tuning, and $y_i$ is the \(i\)-th golden response for \(p\).

\paragraph{Reward Modeling.}
Then, we design the RL's reward to teach our model how to predict the popularity of our generated responses.
Specifically, it takes in a post and response and outputs a scalar reward by comparing
%In our settings, the reward model is trained on a dataset of comparisons 
between two responses given the same post.
The reward difference indicates that one response has more like numbers than the other.
The loss function for the reward model is:

{\small
    \begin{align}
         \mathcal{L}_{RM}(\theta) = -&E_{(p,y_w,y_l)\sim D_{RM}}\nonumber\\
        &[\log(\sigma(r_{\theta}(p,y_w)-r_{\theta}(p,y_l)))],
    \end{align}
    }

\noindent where $\theta$ is the training parameters of reward model,  $r_{\theta}(p,y)$ is the scalar output of the reward model for post $p$ and response $y$, $y_w$ has higher like numbers than $y_l$, and $D_{RM}$ is the reward modeling dataset.

\paragraph{Reinfocement Learning.}
Inspired by InstructGPT's practice, we further update the SFT language model using PPO ~\cite{PPO} to leverage SFT results into the RL framework.
Its loss function can be briefly described as follows:
    \begin{equation}\small
        \mathcal{L}_{RL}(\phi)=-E_{p\sim D_{RL}, y\sim \pi_{\phi}^{RL}(p)} r_{\theta}(p,y)
    \end{equation}

\noindent where $\pi_{\phi}^{RL}$ is the policy RL aims to optimize, which is initialized by the SFT language model. 
Post $p$ is sampled from train dataset $D_{RL}$, $y$ is the output responses of policy given $p$. 
For clarity of presentation, we omit the detail of PPO here and refer readers to \citet{PPO}.
%More details can be found in the PPO more details can be found in PPO.

\subsection{Curriculum Learning-Enhanced PPO}
\label{ssec:cl-ppo}

We can preliminarily align the language model with popularity through the aforementioned learning. 
However, unlike InstructGPT with real human feedback, we use like numbers as automatic labels for assessing response popularity, which is noisy and easily influenced by many factors beyond text. 
% The reward model performance reported in Appendix \ref{reward model} shows it only achieves around 0.6 accuracy in the test set ascribed to the noisy labels.

We thereby incorporate curriculum learning into the PPO algorithm (and present CL-PPO).
It helps filter out noisy training samples while differentiating among training samples' difficulty levels %for optimized learning with noisy labels.
%. It enables effective optimization for the learning pace from easy to hard. 
%We design CL-PPO 
for better-aligning LMs' popularity learning with noisy labels.
CL-PPO has three novel components --- reward enhancement (to provide task-specific supervision), reward ranking (to remove noisy training samples), and self-paced reward sampling (to allow easy-to-hard training) as follows.

\paragraph{Reward Enhancement.}
In reinforcement learning, the rewards not only come from the reward model but also include those directly related to the task \cite{DBLP:journals/corr/abs-2305-05658}, such as the rewards a robotic vacuum cleaner receives for collecting garbage or the rewards earned from finding the exit in a maze game. 
Inspired by this, PopALM integrates a reward enhancement mechanism, using the overlap between the output and highly upvoted responses as a task-specific reward signal.
The reward for a generated response $y$ given post $p$ is defined as:
    \begin{equation}\small
    \label{Eq: reward enhancement}
        r_{\theta}^{e}(p,y) = r_{\theta}(p,y) + \alpha \mathop{\max}_{\hat{y} \in \hat{Y}}(Rouge(y,\hat{y})), 
    \end{equation}
where $r_{\theta}^{e}(p,y)$ is the enhanced reward, $\alpha$ is a weight coefficient, and $\hat{Y}$ is the golden trendy responses. 
$Rouge(y,\hat{y})$ returns the ROUGE-L score between a generated response $y$ and a golden response $\hat{y}$, where the highest ROUGE-L between them is selected to enhance the reward.

\paragraph{Reward Ranking.}
To mitigate the effects of noisy training samples, we introduce a reward ranking mechanism for PPO to increase the training sample quality. 
Specifically, consider a batch of posts, denoted as $\{p_1, p_2, .., p_b\}$ (where $b$ represents the batch size); 
PopALM aims to gain a one-to-many capability to generate multiple trendy responses for each post. 
To that end, for each post in the batch, we generate $m$ responses using a language model with a top-$p$ sampling method \cite{DBLP:conf/iclr/BasuRKV21}.
Then, we obtain the reward $r^{e}_{\theta}$ for each sampled response through the reward model and enhancement mechanism.
Finally, based on $r^{e}_{\theta}$ (reflecting the reward model's confidence), we rank the collected samples and shortlist the $1/k$ percent of samples with the highest reward to engage in the subsequent training. Samples with low rewards are discarded because they signify low prediction confidence and are considered noisy samples.

    \begin{algorithm}[tb]
    \caption{Curriculum Learning-Enhanced PPO\label{algorithm}}\small

    \textbf{Input}: RL Training dataset $D_{RL}$, policy ${\pi}_{\phi}^{RL}$, batch size $b$, reward model $r_{\theta}$, pace parameter $\mu$, acceptance ratio 1/k.
    \begin{algorithmic}[1]
        \FOR{batch $D_b$ from $D_{RL}$}
            \FOR{each $p$ $\in$ \textbf{$D_{b}$}}
                \STATE Predict $m$ trendy responses via top-p sampling, $Y$ = $\{y_{1},y_2,...,y_m\}$ $\sim$ ${\pi}_{\phi}^{RL}$.
                \STATE Compute the reward of each response  $\{r_{\theta}(p,y_1),  r_{\theta}(p,y_2), ..., r_{\theta}(p,y_m)\}$  
                \STATE Compute the enhanced reward using Eq.\ref{Eq: reward enhancement} $\{r_{\theta}^{e}(p,y_1), r_{\theta}^{e}(p,y_2), ..., r_{\theta}^{e}(p,y_m)\}$  
            \ENDFOR
            \STATE Rank reward and select $\lfloor (b*m)/k \rfloor$ training samples with maximum rewards.
            \STATE Select the training samples with higher rewards via self-paced sampling.
            \STATE  Update policy ${\pi}_{\phi}^{RL}$ using Eq.\ref{Eq: cl-ppo}
            \STATE  Update the learning pace via $\lambda \leftarrow \lambda - \mu\lambda$
        
        \ENDFOR
        
    \end{algorithmic}
    \end{algorithm}
\paragraph{Self-paced Sampling.}
With the shortlisted training samples, we further incorporate the self-paced learning method from curriculum learning to enhance learning efficiency.
The intuition is to mimic human knowledge acquisition, starting from simple concepts and gradually tackling more difficult ones requiring advanced skill sets.
Here we measure training samples' learning difficulties with their rewards. Examples with higher rewards have higher prediction confidence, making them easier to learn from.
%we believe that learning from 
We can thus start with the higher-rewarded samples and then move to those with lower rewards. 
%would benefit our overall learning efficiency.
The ultimate learning objective of CL-PPO is defined as follows:
{
\small
\begin{align}
\label{Eq: cl-ppo}
    \mathcal{L}_{CL-PPO}(\phi) &= -E_{(p,y_i)\sim D_{Rank}}\nonumber\\
    &[r_{\theta}^{e}(p,y_i)v_i-\lambda\sum_{i=1}^{|D_{Rank}|}v_i]\nonumber\\
    s.t. \quad v_i =    \quad   &
        \begin{cases}
        1 & \text{if } r_{\theta}^{e}(p,y_i) \geqq \lambda, \\
        0 & otherwise,
        \end{cases}
\end{align}
}
Here $v_i \in \{0, 1\}$ indicates whether the training sample $(p, y_i)$ is selected,
$\lambda$ acts as a threshold to the sampling process and is updated at every training step. 
In detail, for the reward $r_{\theta}^{e}(p,y_i)$ maintained after reward ranking, if it is smaller than the threshold $\lambda$, we set $v_i$ zero as shown in Eq.\ref{Eq: cl-ppo}. 
In this way, during the initial training, responses with larger rewards (corresponding to more popular responses) predominantly contribute to the learning process.
As the training progresses,  $\lambda$ gradually decreases, incorporating lower-rewarded samples to increase the model's generalization capability. Algorithm \ref{algorithm} presents an overview of the entire training process of CL-PPO.

\section{Experimental Setup}
 \begin{table}[!t]
        \centering
        \small
        \begin{tabular}{l|cccc}
        \toprule
            &\textbf{SFT} &\textbf{RM} & \textbf{RL} \\
            \hline
                \textbf{Training}&2,5140&9,985&2,514  \\
                \textbf{Development}   &867&3451&867     \\
                \textbf{Test}   &1,824&7,249&1,824     \\
                \hline
                \textbf{Avg. Posts}  &\multicolumn{3}{c}{119.8}\\
                \textbf{Avg. Responses} &\multicolumn{3}{c}{25.8} \\
        \bottomrule
        \end{tabular}
        \caption{\label{table:data_stat}Statistics of SFT, RM, and RL datasets, followed by the average length (token number) of posts and responses from the raw data.
        %for Parameter-Efficient Fine-tuning, Reward Model training, and our PopALM training.
        }
        \vspace{-1em}
\end{table}

\paragraph{Dataset.}
To set up the experiment, we assembled a new dataset from Weibo, a popular Chinese microblog. 
For data collection, we first obtained the most popular hashtags that have been in use since January 2022, reflecting trending social media events. 
Then, we gathered the raw posts associated with each hashtag using Weibo's search API \footnote{\url{https://open.weibo.com/wiki/C/2/search/statuses/limited}} and selected the post that garnered the most comments as an event description.  
Next, for each selected post, we extracted its comments using the platform's comment API. Finally, our dataset comprised approximately 70,000 posts and 24 million comments filtered from the raw datasets. We did not specifically filter out comments posted by popular authors, even though anything they post might receive many likes. The reason is that many popular authors might also be opinion leaders, often leading the mainstream voice on social media. Furthermore, our model has the capability to filter out some noise responses.
% 
% \footnote{\url{https://open.weibo.com/wiki/2/comments/show}}

Based on the raw data, we gathered three subsets for model training and testing: 
(1) \textit{SFT} dataset (with popular responses) to fine-tune the language model for trendy response prediction; we selected the top 10 comments for each post as the gold response as the reference. 
(2) \textit{RM} dataset (with ranked responses) to train our reward model, where the top 3 comments served as the trendy responses, paired with negative samples of less-liked responses. 
(3) \textit{RL} dataset to train RL's policy to generate responses and provide trendy responses as signals for reward enhancement. Table \ref{table:data_stat} shows these datasets' statistics.
As can be seen, responses are much shorter on average than posts.
It shows that audiences tend to voice their viewpoints concisely, whereas posts may contain richer information for event reporting.

\begin{figure}[!t]
    \centering
    \includegraphics[scale=0.5, trim=0 0 0 0]{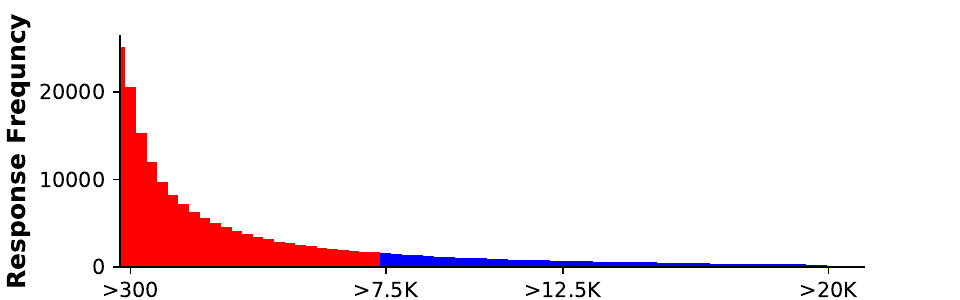}
    \vspace{-0.5em}
    \caption{ \label{fig: response distribution}
    Distribution of response frequency (y-axis) over like numbers (x-axis). 
    Red bars correspond to the top 50\% more popular responses and the rest are blue.
    %X-xis: like numbers. Y-xis: Response frequency. 
    }
    \vspace{-1em}
   
\end{figure}

To further analyze response popularity, we examine the SFT data and display the frequency distribution over like numbers in Figure \ref{fig: response distribution}. 
It is observed that the majority of responses garnered over 300 likes, meaning that our dataset exhibits sufficient samples for learning trendy responses. 
%the popularity of the responses we selected. 
Meanwhile, most responses demonstrate like numbers between 300 to 7,500, whereas the very popular ones (e.g., with over 7,500 likes) appear sparsely.
%responses exceeding 7,500 likes. 
This exhibits a long-tail distribution and challenges our learning to predict trendy responses.
%in the frequency of the overall responses. This distribution poses a 
%challenge to our objective of predicting trendy responses.

\paragraph{Pre-Processing.}
Following common practice \cite{lu-etal-2021-engage}, we first purged the metadata, e.g., the author's information and emoji labels, while substituting links and user mentions (denoted as @username). 
Then, we employed the open-source Jieba toolkit for Chinese word segmentation.
% \footnote{\url{https://github.com/fxsjy/jieba}}

\paragraph{Model Setup.}
Here, we describe how we set our model. 
Based on the statistics of 
%the average length of posts and responses 
in Table \ref{table:data_stat},  we capped the post length to 128 and the response prediction length to 32. 
%in all models. 
To generate diverse responses, we adopt top-$p$ sampling in our experiment with the top-$p$ set to 0.7 and the temperature to 0.95.
For the SFT phrase, we set the learning rate to 0.002 and batch size to 16 for all models. 
We use GPT-2 \cite{gpt2} as the initial reward model. 
For CL-PPO, the weight coefficient $\alpha$ is set to 0.5, the acceptance ratio $k$ is set to 3, the threshold $\theta$ is initialed as 1, and the learning pace $\mu$ is set to 0.2. 
\footnote{Our code and dataset are available at \url{https://github.com/ErxinYu/PopALM}.}
% \begin{table*}[!ht] 
%     \centering
%     \resizebox{\columnwidth*2}{!}{%
%     \begin{tabular}{lcccccc}
%         \hline                

%         % &&&&&&&\multicolumn{2}{c}{A}&\multicolumn{2}{c}{B}&\multicolumn{2}{c}{C}\\
        
%         {\bf Models}& ROUGE-1 & ROUGE-2  & ROUGE-L & BLEU & Distinct-1 & Distinct-2 \\
%         T5   \\
%         Bart  \\
%         ChatGPT \\
%         ChatGLM  \\
%         ChatGLM-Ptuing   \\
%         ChatGLM-PPO   \\
%         {\bf ChatGLM-PPO-Self}\\
%         \hline
%     \end{tabular}
%     }
%     \vspace{-3mm}
%     \caption{Comparisons of performance on six STS datasets. We report average performance for five different random seeds. The better results of different transformer-based model are highlighted in bold (according to the pairwise t-test with 95\% confidence).}  
% \label{overallPeromance}
% \vspace{-5mm}
% \end{table*}  

\begin{table*}[!ht] 
    \centering
    \resizebox{\columnwidth*2}{!}{%
    \begin{tabular}{llcccccccccccc}
        \toprule
        \multirow{2}*{\bf Models} & \multicolumn{4}{c}{Top-1} &\multicolumn{4}{c}{Top-3}&\multicolumn{4}{c}{Top-5}  \\
        \cmidrule(r){2-5} \cmidrule(r){6-9} \cmidrule(r){10-13}
        &R-1&R-2&R-L & BU & R-L & BU & MD-1 & MD-2 & R-L & BU & MD-1 & MD-2\\
        \hline
        &\multicolumn{10}{c}{\bf  Language Models (\textit{w/o SFT})} \\        

        GPT-2   &16.31 &1.79 &11.69 &2.71 &13.17 &3.08 &0.292 &0.483 &14.32 &3.57 &0.228 &0.427   \\
        LLaMA   &1.06  &0.01 &0.85  &0.17 &1.32  &0.29 &0.134 &0.597 &1.669 &0.31 &0.101 &0.567\\
        ChatGLM &14.77 &2.16 &10.88 &3.16 &11.65 &3.45 &0.182 &0.424 &12.19 &3.65 &0.121 &0.320\\
        \hline
        &\multicolumn{10}{c}{\bf  Language Models \textit{(w/ SFT)}} \\
        % GPT-2 (LoRA)    &17.09 &1.35 &12.02 &2.41 
        % &13.49 &2.66&0.264&0.394
        % &13.95&2.81&0.197&0.332\\ 
        DialoGPT &14.22 &1.35 &11.38 &2.11 &12.03 &2.17 &0.143 &0.235 &12.51 &2.27 &0.100 &0.179   \\
        
        CDial-GPT &17.01 &0.79 &12.30 &1.77 &13.13 &1.92 &0.157 &0.223 &13.10 &1.91 &0.068 &0.117   \\

        GPT-2 (P-T)    &18.29 &1.79 &11.69 &2.71 &14.05 &3.31&0.213&0.252&15.15&3.66&0.158&0.214\\
        
        % LLaMA (LoRA)    &16.86 &1.88 &13.30 &3.80 
        % &15.93 &4.56&0.405&0.730
        % &18.00&5.29&0.326&0.664\\
        
        LLaMA (P-T)    &16.87 &1.65 &13.31 &3.27 &16.05 &4.14&0.450&0.755&17.55&4.60&0.369&0.703\\

        ChatGLM (LoRA) &18.39 &3.11 &15.08 &5.72 &19.50 &7.84&0.489&0.590&21.70&8.82&0.382&0.497\\
        ChatGLM (P-T)  &18.63 &3.29 &15.94 &6.16 &19.69 &7.79&0.498&0.576&22.98&9.38&0.431&0.501\\

        \hline
        &\multicolumn{10}{c}{\bf Popularity-Aligned Language Models (PopALM)} \\
        ChatGLM (PPO) &18.61 &3.09  &16.06 &6.19 &20.01 &7.91&0.511&0.583&22.66&9.27&0.437&0.506\\
        {\bf PopALM} &{\bf 19.49} &{\bf 3.69}  &{\bf 16.42} &{\bf 6.35} &{\bf 21.50} &{\bf 8.43} &{\bf 0.541} &{\bf 0.632} &{\bf 23.58} &{\bf 9.63} &{\bf 0.452} &{\bf 0.511}\\

        \bottomrule
        
    \end{tabular}
    }
    \caption{We present the automatic evaluation results for the top-1, top-3, and top-5 trendy responses predicted by PopALM, i.e., \textbf{ChatGLM (CL-PPO)}. 
    For the top-1 prediction, we report the performance metrics R-1 (ROUGE-1), R-2 (ROUGE-2), R-L (ROUGE-L), and BU (BLEU). For top-3 and top-5 predictions, we provide R-L and BU to measure the overlap performance and employ MD-1 (M-Distinct-1) and MD-2 (M-Distinct-2) to evaluate the diversity performance. We report the average performance for five different random seeds, and the better results (compared to PPO) are highlighted in bold, indicating a statistically significant difference ($p$ < 0.05) from baselines with bootstrap resampling \cite{koehn-2004-statistical}.}
\label{table:auto eval}
\vspace{-3mm}
\end{table*}  
\paragraph{Evaluation Metrics.}
For \textit{Automatic Evaluation}, we follow \citet{zhang-etal-2020-dialogpt} to compare output and gold responses and evaluate the output quality with overlapping-based metrics ROUGE ~\cite{lin-2004-rouge} and BLEU \cite{papineni-etal-2002-bleu} scores. 
Besides, we use M-Distinct-n ~\cite{li-etal-2016-persona} to score the diversity of responses, which measures the model's ability to generate multiple diverse responses for the same test posts. 
% In particular, a post has multiple trendy response references in our dataset, we thus compare the generated response with each reference and report the highest score.

For \textit{Human Evaluations}, we invited human raters with NLP backgrounds to rate the generated responses on a 5-point Likert scale on the following dimensions. 
%three dimensions: \textit{Relevance}, \textit{Specification}, and \textit{Popularity}. 
%\textit{Relevance} reflects the output's consistency to the post's context.
%whether the response is about the main story of the posts, one side part of the posts, or irrelevant to the posts. 
\textit{Informativeness} reflects how much information is presented in the generated results. 
\textit{Specification} assesses the degree of the output containing specific viewpoints.
\textit{Popularity} measures the potential of the response to be liked by many users and become popular.
In addition, we involved an \textit{Overall} score to reflect raters' general feelings by combining the above three dimensions.
%whether the generated responses will likely receive likes 
%or approval from the readers on social media.
%Each aspect is scored on a scale from 1 to 5. 
Here, we randomly select 100 posts from the test set and enlist raters to assess the responses without knowing which model generated them.

\paragraph{Comparison Setup.}

For the pre-trained models, we adopt several language models that have not been fine-tuned on our dataset: 
1) \underline{GPT-2} \cite{gpt2} is a decoder-based language model for generating contextually relevant and coherent text. 
2) \underline{DialoGPT} \cite{zhang-etal-2020-dialogpt} is a response generation model based on GPT-2, pre-trained on a large corpus of social media text. 
3) \underline{CDial-GPT} \cite{wang2020chinese} is first pre-trained on a Chinese novel dataset and then post-trained on a large-scale Chinese dialog dataset, demonstrating strong response generation capabilities.
4) \underline{LLaMA} \cite{touvron2023llama} is a foundational large language model designed for researchers. 
5) \underline{ChatGLM} is an open bilingual language model based on the General Language Model \cite{du-etal-2022-glm}. 
% 6) \underline{ChatGPT}  is a sibling model to InstructGPT \cite{ChatGPT}, which is trained to follow human instructions and achieve state-of-the-art performance in most natural language tasks.

For DialoGPT and CDial-GPT, we employ full-parameter fine-tuning on our dataset.
For other models under SFT and PPO settings, to enable efficient adaptation of pre-trained language models to our task, we employ two Parameter-Efficient Fine-Tuning (PEFT) methods: 
1) \underline{P-Tuning (P-T)} \cite{liu-etal-2022-p} tunes continuous prompts with a frozen language model. 
2) \underline{LoRA} \cite{lora} injects trainable rank decomposition matrices into the Transformer.

\section{Experimental Results}
\label{sec:exp-results}

This section first discusses the main comparison results in Section \ref{ssec:main-results}, followed by the ablation study to examine the varying CL-PPO strategies' contributions in Section \ref{ssec:ablation-study}.
Then, we quantify the effects of language models, PEFT methods, and training data scales in Section \ref{ssec:quantitative-analysis}.
After that, we qualitatively analyze why PopALM can exhibit superior results through a case study in Section \ref{ssec:case-study}.
Finally, we demonstrate the impact of generated responses in Section \ref{ssec:other tasks}.

\subsection{Main Comparison Results\label{ssec:main-results}}

\paragraph{Automatic Evaluation Results.}
Table \ref{table:auto eval} shows the result, where we draw the following observations: 
1) The previous response generation models, DialoGPT and CDial-GPT, despite being trained on large-scale conversational text, still fall short in predicting popular responses.
2) Compared to the original language models, PEFT allows models to yield better responses. 
This suggests that only training a minor fraction of parameters can also equip language models with the capability to predict popularity.
3) Using the PPO method to align the language model with popularity is beneficial. 
However, some metrics are decreased after the PPO training, possibly due to the negative effects of noisy labels. 
4) Our proposed PopALM significantly outperforms the PPO in all automatic metrics. 
Moreover, in the top-3 and top-5 predictions, the responses produced by CL-PPO exhibit greater diversity.
The above results suggest the effectiveness of CL-PPO in
%our model in sampling high-quality data, thus 
mitigating the issue of noisy labels and allowing more efficient learning for trendy response prediction. 
%within trendy response prediction.
\begin{table}[!tb] 
    \resizebox{\columnwidth}{!}{%
    \begin{tabular}{lcccccccc}
        \toprule               
        {\bf Models}&Info &Spec &Pop &Overall \\
        \hline 
        % ChatGPT &{\bf 3.38}&2.64&2.33&2.59\\
        ChatGLM &\textbf{2.12}&1.70&1.75&1.86\\
        ChatGLM(P-T) &1.65&2.92&2.11&2.23\\
        ChatGLM(PPO) &1.73&2.89&2.26&2.43\\
        {\bf PopALM } &1.91&{\bf 3.14}&{\bf 2.89}&{\bf 2.65}   \\
        \bottomrule
    \end{tabular}
    }
    \caption{\label{table:humanEvaluation}Human Evaluation on randomly sampled 100 test samples. We compare ChatGLM with P-T/PPO, and PopALM model.}  

\end{table}  

\begin{table}[!tb] 
    \resizebox{\columnwidth}{!}{%
    \begin{tabular}{llccccc}
        \toprule
        \multirow{2}*{\bf Models} & \multicolumn{2}{c}{GPT-2} &\multicolumn{2}{c}{LLaMA}&\multicolumn{2}{c}{ChatGLM}  \\
        \cmidrule(r){2-3} \cmidrule(r){4-5} \cmidrule(r){6-7}
        &LoRA&P-T&LoRA&P-T&LoRA&P-T\\
        \hline
        PPO &13.79 &14.77 &16.21 &16.13&19.89&20.01 \\
        {\bf CL-PPO}&{\bf 14.98} &{\bf 15.56}&{\bf 17.54} &{\bf 17.23}&{\bf 20.77}&{\bf 21.50} \\

        \bottomrule
        
    \end{tabular}
    }
    \caption{Result of top-3 prediction ROUGE-L score with varying Language Models (LMs) with PEFT.}  
\label{table:lm_peft_effect}
\vspace{-3mm}
\end{table}  
\paragraph{Human Evaluation.}

We select PopALM and three variants of ChatGLM to compare how human readers evaluate their output.
The results are shown in Table \ref{table:humanEvaluation}. PopALM gains higher scores in specification and popularity, while its performance falls on the informativeness metric compared to ChatGLM. 
It may be that PopALM generates more specific responses, thereby losing some general information. 
Meanwhile, the responses generated by PopALM are more stylized towards social media than the other two fine-tuning methods of ChatGLM.
The result shows that through popularity-aligned reinforcement learning, language models yield more specific points to reflect the public's concerns and are more likely to receive likes.

\subsection{Ablation Study\label{ssec:ablation-study}}

The above results show the overall superiority of CL-PPO. 
To further investigate the effects of its components, we conduct an ablation study with the results displayed in Figure \ref{ablation}.
As can be seen, our three proposed components all contribute positively across different language models (ChatGLM, LLaMA, and GPT-2) and PEFT methods (LoRA, P-tuning). 
In particular, self-paced sampling contributes substantially when ChatGLM is used as the backbone language model. The performance drops by 1.48 and 1.11, respectively, and even falls below PPO's when self-paced sampling is reduced. 
This illustrates that prioritizing high-reward examples for early learning is beneficial for the models to learn trendy response prediction efficiently.

\subsection{Quantitative Analysis}\label{ssec:quantitative-analysis}
We then quantify PopALM with varying training setups to deepen the understanding of it.

\paragraph{Varying Language Models and PEFT Methods.}
We first investigate the backbone language models (LMs) and PEFT methods and display the results in Table \ref{table:lm_peft_effect}.
It shows that CL-PPO exhibits improved performance over the original PPO across different combinations of LMs and PEFT methods. This validates our model as a plug-and-play approach that can be effectively applied to various LMs.

\begin{figure}[!t]
    \centering
    \includegraphics[width=0.5\textwidth]{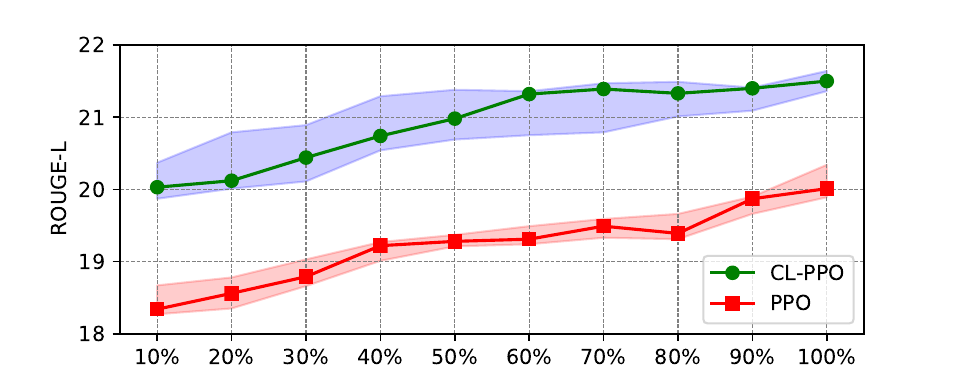}
    \caption{Effects of training data scales (x-axis). The y-axis shows the ROUGE-L score of the top-3 prediction based on ChatGLM. The colored bands
indicate \(\pm\)1 standard deviation corresponding to different percentages of training data.}
    \label{data_effect}
    \vspace{-5mm}
\end{figure}
        \begin{figure*}
        \centering
        \subfigure[GPT-2]{
            \begin{minipage}[t]{0.33\textwidth}
            \centering
            \includegraphics[width=1\textwidth]{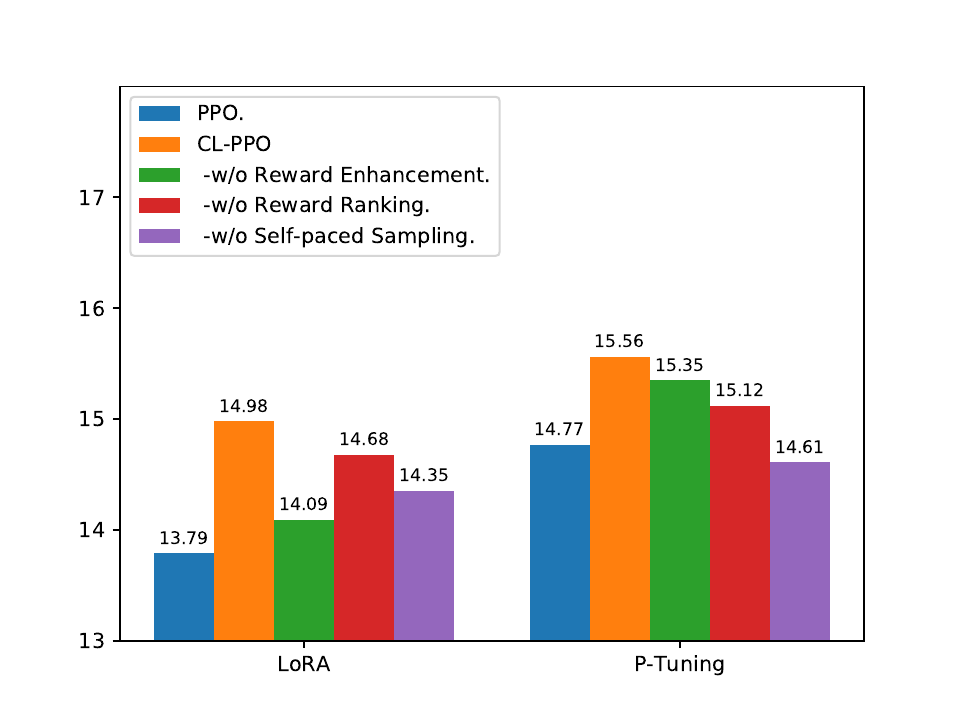}
            \vspace{-5mm}
            \end{minipage}%
            }%
        \subfigure[LLaMA]{
            \begin{minipage}[t]{0.33\textwidth}
            \centering
            \includegraphics[width=1\textwidth, trim = 0 0 0 10]{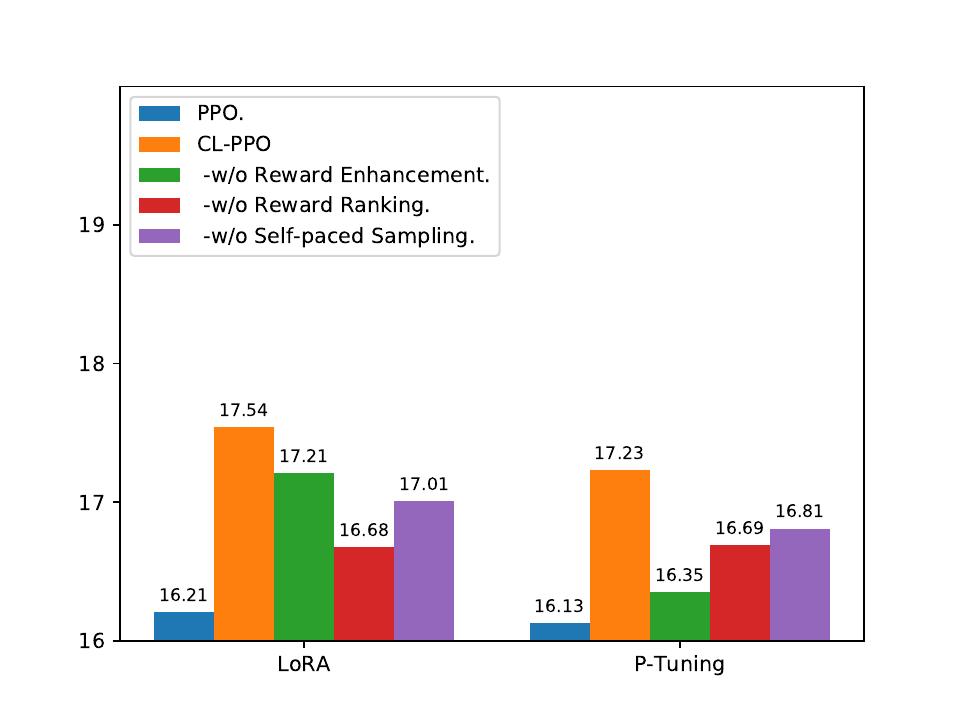}
            \vspace{-5mm}
        \end{minipage}%      
        }%
        \subfigure[ChatGLM]{
            \begin{minipage}[t]{0.33\textwidth}
            \centering
            \includegraphics[width=1\textwidth, trim = 0 0 0 10]{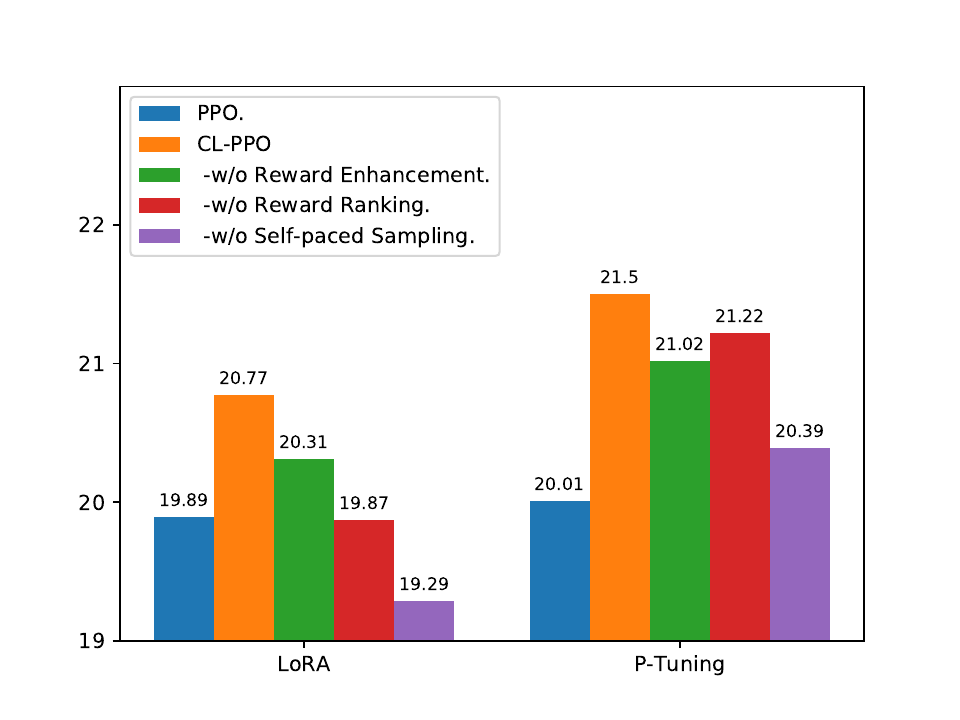}
            \vspace{-5mm}
        \end{minipage}%      
        }%
        \centering
        \vspace{-3mm}
        \caption{Ablation study on CL-PPO. We report the ROUGE-L scores of the Top-3 trendy response predictions for GPT-2, LLaMA, and ChatGLM. For them each, we show PEFT results of LoRA on the left and P-Tuning on the right. For each barplot, the bars from left to right show PPO, CL-PPO, followed by the CL-PPO ablations w/o Reward Enhancement, w/o Reward Ranking, and w/o Self-paced Sampling.}
        \vspace{-3mm}
        \label{ablation}
        \end{figure*}

\paragraph{Varying Training Data Scales.}
We then test PopALM's sensitivity to training data scales by training it with different data percentages. 
As shown in Figure 5, our proposed CL-PPO training algorithm consistently outperforms the original PPO regardless of the volume of training data used, ranging from 10\% to 100\%. This suggests the stable and consistent performance of CL-PPO across different training data amounts.

\subsection{Case Study\label{ssec:case-study} }

We exemplify the case in Figure \ref{fig:intro-case} and compare different models' output in Table \ref{table:caseStudyEn} to qualitatively analyze why PopALM can yield better results.
%we compare responses generated by different models
Recall that the post describes an event in which a volunteer leaked
exam questions and posted them online, and the school explained that the volunteer did not have full access to the exam paper. 

The output of ChatGPT (see Figure \ref{fig:intro-case}) and untrained ChatGLM are more akin to summarizing the post, tending to be more generic. 
After fine-tuning, ChatGLM (P-T) can generate responses in a social media style, incorporating its own opinion. After PPO training, the model chose to train on responses with higher scores in the reward model. However, due to the noisy labels, the scores given by the reward model may not be entirely accurate, misleading the generation results, as shown in the table. 
In contrast, CL-PPO allows more effective training with noisy labels and consequently better aligns the output with trendy responses.

\begin{CJK}{UTF8}{gbsn} 
\begin{table}[!t] 
    \resizebox{\columnwidth}{!}{%
    {\small
    \begin{tabular}{p{\columnwidth}}
        \toprule
         {\bf ChatGLM:} \ Volunteer pasted barcodes on exam  bag  and took photos, ruling out contact with 
          papers.  Public concern may arise due to planned higher authority investigations. \\
         {\bf ChatGLM (P-T):} \ Don't even bring up volunteers anymore. \\
         {\bf ChatGLM (PPO):} \ Volunteers gonna volunteer, candidates gonna candidate, it's just pointless. \\
        {\bf PopALM }: \ Volunteers need training. Fair exams should be a guarantee! \\
        \bottomrule
    \end{tabular}
    }
}
    \caption{Case study of different models' output for the post in Figure \ref{fig:intro-case}.
    }  
    % Here we show the English translation and put the original content in Chinese in Appendix \ref{case-study appendix-Chinese}.
\label{table:caseStudyEn}
\vspace{-3mm}
\end{table}  
\end{CJK}

% \begin{CJK}{UTF8}{gbsn} 
% \begin{table}[!t] 
%     \resizebox{\columnwidth}{!}{%
%     {\small
%     \begin{tabular}{p{\columnwidth}}
%         \toprule
%          {\bf ChatGLM:} \ 志愿者在考试袋上粘了条形码并拍了照，没接触试卷哦～ 高层有计划调查，大家注意，可能会引起公众关注哦。(\textit{Volunteer pasted barcodes on exam  bag  and took photos, ruling out contact with 
%           papers.  Public concern may arise due to planned higher authority investigations}.) \\
%          {\bf ChatGLM (P-T):} \ 不要再提志愿者了。(\textit{Don't even bring up volunteers anymore.}) \\
%          {\bf ChatGLM (PPO):} \ 志愿者会做志愿，候选人会参选，这真的没意义。(\textit{Volunteers gonna volunteer, candidates gonna candidate, it's just pointless.}) \\
%         {\bf PopALM }: \ 志愿者需要培训, 保证公平考试! (\textit{Volunteers need training. Fair exams should be a guarantee!}) \\
%         \bottomrule
%     \end{tabular}
%     }
% }
%     \caption{Case study of different models' output for the post in Figure \ref{fig:intro-case}.
%     }  
%     % Here we show the English translation and put the original content in Chinese in Appendix \ref{case-study appendix-Chinese}.
% \label{table:caseStudyEn}
% \vspace{-3mm}
% \end{table}  
% \end{CJK}

\subsection{Impact of Generated Response\label{ssec:other tasks} }

In Table \ref{table:impact_responses}, we demonstrate the impact of the responses generated by our model on two tasks: poll question generation \citep{lu-etal-2021-engage} and social emotion prediction \citep{ding-etal-2020-hashtags}. Poll question generation aims to automatically generate questions for posts, in which popular responses can reflect the public's concerns and engage them in discussions. Social emotion prediction involves predicting the public's attitude towards posts. Including mainstream reactions can help assess the general attitude.

We tested the poll question generation task based on ChatGLM and set up a comparative experiment: one approach is to input only the post to generate a poll question, while the other concatenates the post and responses as input. We employed RoBERTa \cite{roberta} as the classifier for the social emotion prediction task and adopted the same comparative experiment. As can be seen from the table, incorporating PopALM-generated responses yields better results for both tasks. However, using responses directly generated by ChatGLM doesn't have much effect.
 Moreover, the results indicate that the PopALM-generated responses could perform comparable to real responses.
\begin{table}[!tb] 
    \resizebox{\columnwidth}{!}{%
    \begin{tabular}{llcccc}
        \toprule
        \multirow{2}{*}{\bf Methods} & \multicolumn{2}{c}{PQG} &\multicolumn{2}{c}{SEP} \\
        \cmidrule(r){2-3} \cmidrule(r){4-5}
        & $R-1$ & $R-L$ & $F1_{macro}$ & $F1_{micro}$ \\
        \midrule
       \textbf{W/O Responses} & 0.331 & 0.305 & 0.312 & 0.408 \\
       \textbf{W/ ChatGLM Responses} & 0.323 &0.314 & 0.303 & 0.401 \\
        \textbf{W/ PopALM Responses} & 0.363 & \bf{0.337} & 0.322 & 0.422 \\
        \textbf{W/ Real Responses} &  \textbf{0.367} & 0.331 &  \textbf{0.325} & \textbf{0.426} \\
        \bottomrule
    \end{tabular}
    }
    \caption{Performance of the different responses on the Poll Question Generation (PQG) and Social Emotion Prediction (SEP) tasks. We use Rouge-1 and Rouge-L to evaluate PQG, and macro F1 and micro F1 to assess SEP.}
    \label{table:impact_responses}
    \vspace{-3mm}
\end{table}

% \begin{table}[!tb]
%     \centering
%     \begin{tabular}{llcccc}
%         \toprule
%         \multirow{2}{*}{\textbf{Models}} & \multicolumn{2}{c}{PQG} &\multicolumn{2}{c}{SEC} \\
%         \cmidrule(r){2-3} \cmidrule(r){4-5}
%         & R1 & RL & $F1_{macro}$ & $F1_{micro}$ \\
%         \midrule
%         W/O Comments & 0.353 & 0.319 & 0.312 & 0.408 \\
%          % Apply the gray color to this row
%         Generated Comments & 0.363 & 0.326 & 0.322 & 0.412 \\
%         Real Comments &  0.367 & 0.331 &  0.325 &  0.426  \\
%         \bottomrule
%     \end{tabular}
%     \caption{Comparison of models with and without comments on PQG and SEC metrics.}
%     \label{table:lm_peft_effect}
% \end{table}

We study trendy response prediction to predict the mainstream public reaction before an event happens or in its early stages. Beyond the above response-augmented tasks, it also offers other potential applications.
For example, it can be applied in early event analysis to foresee the future impact of a breaking event before many people engage in related discussions. 
Social scientists can also employ our model to simulate the public responses to some social events even though they have not yet happened. Moreover, our study can potentially benefit general comment generation applications \cite{autoCommentGen1,autoCommentGenAAAI,autoGenTOMM} and encourage better user engagement.

\section{Conclusion}
We have presented a study on trendy response prediction for social media events, an area that previously lacked exploration. 
A novel popularity-aligned language model was proposed by integrating a specifically designed curriculum learning strategy into proximal policy optimization to learn popularity from noisy user-like labels.
A large-scale benchmark was constructed, and its experimental results show that PopALM exhibits performance gains to LMs with various training setups.
%to learn popularity and help predict trendy responses. 

\section*{Ethics Statement}

Our paper presents a large-scale Weibo corpus for trendy response prediction, and it will not pose ethical problems. Firstly, these posts and responses are open to the public. Secondly, Weibo allows any user to report suspicious cases that may involve ethical issues, and the reported content will be immediately removed. Finally, the data was gathered via standard data acquisition procedures regulated by Weibo's API and was downloaded exclusively for academic research purposes. 

For our experiments, we have taken steps to anonymize the data to protect privacy, such as removing author names and changing mentions and URL links to generic tags. 
\section*{Acknowledgements}
This work is supported by the NSFC Young Scientists Fund (Project No. 62006203), a grant from the Research Grants Council of the Hong Kong Special Administrative Region, China (Project No. PolyU/25200821), the Innovation and Technology Fund (Project No. PRP/047/22FX), and PolyU Internal Fund from RC-DSAI (Project No. 1-CE1E).
\section{Bibliographical References}\label{sec:reference}
\bibliographystyle{lrec-coling2024-natbib}
\bibliography{lrec-coling2024-example, anthology}

\begin{thebibliography}{40}
\expandafter\ifx\csname natexlab\endcsname\relax\def\natexlab#1{#1}\fi

\bibitem[{Bao et~al.(2020)Bao, He, Wang, Wu, and Wang}]{bao-etal-2020-plato}
Siqi Bao, Huang He, Fan Wang, Hua Wu, and Haifeng Wang. 2020.
\newblock \href {https://doi.org/10.18653/v1/2020.acl-main.9} {{PLATO}:
  Pre-trained dialogue generation model with discrete latent variable}.
\newblock In \emph{Proceedings of the 58th Annual Meeting of the Association
  for Computational Linguistics}, pages 85--96, Online. Association for
  Computational Linguistics.

\bibitem[{Basu et~al.(2021)Basu, Ramachandran, Keskar, and
  Varshney}]{DBLP:conf/iclr/BasuRKV21}
Sourya Basu, Govardana~Sachitanandam Ramachandran, Nitish~Shirish Keskar, and
  Lav~R. Varshney. 2021.
\newblock \href {https://openreview.net/forum?id=W1G1JZEIy5\_} {Mirostat: a
  neural text decoding algorithm that directly controls perplexity}.
\newblock In \emph{9th International Conference on Learning Representations,
  {ICLR} 2021, Virtual Event, Austria, May 3-7, 2021}. OpenReview.net.

\bibitem[{Bengio et~al.(2009)Bengio, Louradour, Collobert, and
  Weston}]{CurriculumLearningICML2009}
Yoshua Bengio, J\'{e}r\^{o}me Louradour, Ronan Collobert, and Jason Weston.
  2009.
\newblock \href {https://doi.org/10.1145/1553374.1553380} {Curriculum
  learning}.
\newblock In \emph{Proceedings of the 26th Annual International Conference on
  Machine Learning}, ICML '09, page 41–48, New York, NY, USA. Association for
  Computing Machinery.

\bibitem[{Brown et~al.(2020)Brown, Mann, Ryder, Subbiah, Kaplan, Dhariwal,
  Neelakantan, Shyam, Sastry, Askell, Agarwal, Herbert-Voss, Krueger, Henighan,
  Child, Ramesh, Ziegler, Wu, Winter, Hesse, Chen, Sigler, Litwin, Gray, Chess,
  Clark, Berner, McCandlish, Radford, Sutskever, and Amodei}]{gpt3}
Tom~B. Brown, Benjamin Mann, Nick Ryder, Melanie Subbiah, Jared Kaplan,
  Prafulla Dhariwal, Arvind Neelakantan, Pranav Shyam, Girish Sastry, Amanda
  Askell, Sandhini Agarwal, Ariel Herbert-Voss, Gretchen Krueger, Tom Henighan,
  Rewon Child, Aditya Ramesh, Daniel~M. Ziegler, Jeffrey Wu, Clemens Winter,
  Christopher Hesse, Mark Chen, Eric Sigler, Mateusz Litwin, Scott Gray,
  Benjamin Chess, Jack Clark, Christopher Berner, Sam McCandlish, Alec Radford,
  Ilya Sutskever, and Dario Amodei. 2020.
\newblock \href {http://arxiv.org/abs/2005.14165} {Language models are few-shot
  learners}.

\bibitem[{Ding et~al.(2020)Ding, Li, and Zhang}]{ding-etal-2020-hashtags}
Keyang Ding, Jing Li, and Yuji Zhang. 2020.
\newblock \href {https://doi.org/10.18653/v1/2020.emnlp-main.106} {Hashtags,
  emotions, and comments: A large-scale dataset to understand fine-grained
  social emotions to online topics}.
\newblock In \emph{Proceedings of the 2020 Conference on Empirical Methods in
  Natural Language Processing (EMNLP)}, pages 1376--1382, Online. Association
  for Computational Linguistics.

\bibitem[{Du et~al.(2022)Du, Qian, Liu, Ding, Qiu, Yang, and
  Tang}]{du-etal-2022-glm}
Zhengxiao Du, Yujie Qian, Xiao Liu, Ming Ding, Jiezhong Qiu, Zhilin Yang, and
  Jie Tang. 2022.
\newblock \href {https://doi.org/10.18653/v1/2022.acl-long.26} {{GLM}: General
  language model pretraining with autoregressive blank infilling}.
\newblock In \emph{Proceedings of the 60th Annual Meeting of the Association
  for Computational Linguistics (Volume 1: Long Papers)}, pages 320--335,
  Dublin, Ireland. Association for Computational Linguistics.

\bibitem[{Gao et~al.(2020)Gao, Zhang, Galley, Brockett, and
  Dolan}]{gao-etal-2020-dialogue}
Xiang Gao, Yizhe Zhang, Michel Galley, Chris Brockett, and Bill Dolan. 2020.
\newblock \href {https://doi.org/10.18653/v1/2020.emnlp-main.28} {Dialogue
  response ranking training with large-scale human feedback data}.
\newblock In \emph{Proceedings of the 2020 Conference on Empirical Methods in
  Natural Language Processing (EMNLP)}, pages 386--395, Online. Association for
  Computational Linguistics.

\bibitem[{Hu et~al.(2022)Hu, Shen, Wallis, Allen{-}Zhu, Li, Wang, Wang, and
  Chen}]{lora}
Edward~J. Hu, Yelong Shen, Phillip Wallis, Zeyuan Allen{-}Zhu, Yuanzhi Li,
  Shean Wang, Lu~Wang, and Weizhu Chen. 2022.
\newblock \href {https://openreview.net/forum?id=nZeVKeeFYf9} {Lora: Low-rank
  adaptation of large language models}.
\newblock In \emph{The Tenth International Conference on Learning
  Representations, {ICLR} 2022, Virtual Event, April 25-29, 2022}.
  OpenReview.net.

\bibitem[{Kano et~al.(2018)Kano, Miura, Taniguchi, Chen, Chen, and
  Ohkuma}]{kano-etal-2018-harnessing}
Ryuji Kano, Yasuhide Miura, Motoki Taniguchi, Yan-Ying Chen, Francine Chen, and
  Tomoko Ohkuma. 2018.
\newblock \href {https://doi.org/10.18653/v1/D18-1144} {Harnessing popularity
  in social media for extractive summarization of online conversations}.
\newblock In \emph{Proceedings of the 2018 Conference on Empirical Methods in
  Natural Language Processing}, pages 1139--1145, Brussels, Belgium.
  Association for Computational Linguistics.

\bibitem[{Koehn(2004)}]{koehn-2004-statistical}
Philipp Koehn. 2004.
\newblock \href {https://aclanthology.org/W04-3250} {Statistical significance
  tests for machine translation evaluation}.
\newblock In \emph{Proceedings of the 2004 Conference on Empirical Methods in
  Natural Language Processing}, pages 388--395, Barcelona, Spain. Association
  for Computational Linguistics.

\bibitem[{Lamprinidis et~al.(2018)Lamprinidis, Hardt, and
  Hovy}]{lamprinidis-etal-2018-predicting}
Sotiris Lamprinidis, Daniel Hardt, and Dirk Hovy. 2018.
\newblock \href {https://doi.org/10.18653/v1/D18-1068} {Predicting news
  headline popularity with syntactic and semantic knowledge using multi-task
  learning}.
\newblock In \emph{Proceedings of the 2018 Conference on Empirical Methods in
  Natural Language Processing}, pages 659--664, Brussels, Belgium. Association
  for Computational Linguistics.

\bibitem[{Lewis et~al.(2020)Lewis, Liu, Goyal, Ghazvininejad, Mohamed, Levy,
  Stoyanov, and Zettlemoyer}]{lewis-etal-2020-bart}
Mike Lewis, Yinhan Liu, Naman Goyal, Marjan Ghazvininejad, Abdelrahman Mohamed,
  Omer Levy, Veselin Stoyanov, and Luke Zettlemoyer. 2020.
\newblock \href {https://doi.org/10.18653/v1/2020.acl-main.703} {{BART}:
  Denoising sequence-to-sequence pre-training for natural language generation,
  translation, and comprehension}.
\newblock In \emph{Proceedings of the 58th Annual Meeting of the Association
  for Computational Linguistics}, pages 7871--7880, Online. Association for
  Computational Linguistics.

\bibitem[{Li et~al.(2016)Li, Galley, Brockett, Spithourakis, Gao, and
  Dolan}]{li-etal-2016-persona}
Jiwei Li, Michel Galley, Chris Brockett, Georgios Spithourakis, Jianfeng Gao,
  and Bill Dolan. 2016.
\newblock \href {https://doi.org/10.18653/v1/P16-1094} {A persona-based neural
  conversation model}.
\newblock In \emph{Proceedings of the 54th Annual Meeting of the Association
  for Computational Linguistics (Volume 1: Long Papers)}, pages 994--1003,
  Berlin, Germany. Association for Computational Linguistics.

\bibitem[{Lin(2004)}]{lin-2004-rouge}
Chin-Yew Lin. 2004.
\newblock \href {https://aclanthology.org/W04-1013} {{ROUGE}: A package for
  automatic evaluation of summaries}.
\newblock In \emph{Text Summarization Branches Out}, pages 74--81, Barcelona,
  Spain. Association for Computational Linguistics.

\bibitem[{Liu et~al.(2022)Liu, Ji, Fu, Tam, Du, Yang, and
  Tang}]{liu-etal-2022-p}
Xiao Liu, Kaixuan Ji, Yicheng Fu, Weng Tam, Zhengxiao Du, Zhilin Yang, and Jie
  Tang. 2022.
\newblock \href {https://doi.org/10.18653/v1/2022.acl-short.8} {{P}-tuning:
  Prompt tuning can be comparable to fine-tuning across scales and tasks}.
\newblock In \emph{Proceedings of the 60th Annual Meeting of the Association
  for Computational Linguistics (Volume 2: Short Papers)}, pages 61--68,
  Dublin, Ireland. Association for Computational Linguistics.

\bibitem[{Liu et~al.(2019)Liu, Ott, Goyal, Du, Joshi, Chen, Levy, Lewis,
  Zettlemoyer, and Stoyanov}]{roberta}
Yinhan Liu, Myle Ott, Naman Goyal, Jingfei Du, Mandar Joshi, Danqi Chen, Omer
  Levy, Mike Lewis, Luke Zettlemoyer, and Veselin Stoyanov. 2019.
\newblock \href {http://arxiv.org/abs/1907.11692} {Roberta: {A} robustly
  optimized {BERT} pretraining approach}.
\newblock \emph{CoRR}, abs/1907.11692.

\bibitem[{Lu et~al.(2021)Lu, Ding, Zhang, Li, Peng, and
  Liu}]{lu-etal-2021-engage}
Zexin Lu, Keyang Ding, Yuji Zhang, Jing Li, Baolin Peng, and Lemao Liu. 2021.
\newblock \href {https://doi.org/10.18653/v1/2021.acl-long.3} {Engage the
  public: Poll question generation for social media posts}.
\newblock In \emph{Proceedings of the 59th Annual Meeting of the Association
  for Computational Linguistics and the 11th International Joint Conference on
  Natural Language Processing (Volume 1: Long Papers)}, pages 29--40, Online.
  Association for Computational Linguistics.

\bibitem[{Ouyang et~al.(2022{\natexlab{a}})Ouyang, Wu, Jiang, Almeida,
  Wainwright, Mishkin, Zhang, Agarwal, Slama, Ray, Schulman, Hilton, Kelton,
  Miller, Simens, Askell, Welinder, Christiano, Leike, and
  Lowe}]{learningFromHumanFeedback}
Long Ouyang, Jeff Wu, Xu~Jiang, Diogo Almeida, Carroll~L. Wainwright, Pamela
  Mishkin, Chong Zhang, Sandhini Agarwal, Katarina Slama, Alex Ray, John
  Schulman, Jacob Hilton, Fraser Kelton, Luke~E. Miller, Maddie Simens, Amanda
  Askell, Peter Welinder, Paul~Francis Christiano, Jan Leike, and Ryan~J. Lowe.
  2022{\natexlab{a}}.
\newblock Training language models to follow instructions with human feedback.
\newblock \emph{ArXiv}, abs/2203.02155.

\bibitem[{Ouyang et~al.(2022{\natexlab{b}})Ouyang, Wu, Jiang, Almeida,
  Wainwright, Mishkin, Zhang, Agarwal, Slama, Ray, Schulman, Hilton, Kelton,
  Miller, Simens, Askell, Welinder, Christiano, Leike, and
  Lowe}]{instructGPTNIPS2022}
Long Ouyang, Jeffrey Wu, Xu~Jiang, Diogo Almeida, Carroll~L. Wainwright, Pamela
  Mishkin, Chong Zhang, Sandhini Agarwal, Katarina Slama, Alex Ray, John
  Schulman, Jacob Hilton, Fraser Kelton, Luke Miller, Maddie Simens, Amanda
  Askell, Peter Welinder, Paul~F. Christiano, Jan Leike, and Ryan Lowe.
  2022{\natexlab{b}}.
\newblock \href
  {http://papers.nips.cc/paper\_files/paper/2022/hash/b1efde53be364a73914f58805a001731-Abstract-Conference.html}
  {Training language models to follow instructions with human feedback}.
\newblock In \emph{NeurIPS}.

\bibitem[{Papineni et~al.(2002)Papineni, Roukos, Ward, and
  Zhu}]{papineni-etal-2002-bleu}
Kishore Papineni, Salim Roukos, Todd Ward, and Wei-Jing Zhu. 2002.
\newblock \href {https://doi.org/10.3115/1073083.1073135} {{B}leu: a method for
  automatic evaluation of machine translation}.
\newblock In \emph{Proceedings of the 40th Annual Meeting of the Association
  for Computational Linguistics}, pages 311--318, Philadelphia, Pennsylvania,
  USA. Association for Computational Linguistics.

\bibitem[{Qin et~al.(2018)Qin, Liu, Bi, Wang, Liu, Hu, Zhao, and
  Shi}]{qin-etal-2018-automatic}
Lianhui Qin, Lemao Liu, Wei Bi, Yan Wang, Xiaojiang Liu, Zhiting Hu, Hai Zhao,
  and Shuming Shi. 2018.
\newblock \href {https://doi.org/10.18653/v1/P18-2025} {Automatic article
  commenting: the task and dataset}.
\newblock In \emph{Proceedings of the 56th Annual Meeting of the Association
  for Computational Linguistics (Volume 2: Short Papers)}, pages 151--156,
  Melbourne, Australia. Association for Computational Linguistics.

\bibitem[{Radford and Narasimhan(2018)}]{gpt1}
Alec Radford and Karthik Narasimhan. 2018.
\newblock Improving language understanding by generative pre-training.

\bibitem[{Radford et~al.(2019)Radford, Wu, Child, Luan, Amodei, and
  Sutskever}]{gpt2}
Alec Radford, Jeff Wu, Rewon Child, David Luan, Dario Amodei, and Ilya
  Sutskever. 2019.
\newblock Language models are unsupervised multitask learners.

\bibitem[{Raffel et~al.(2020)Raffel, Shazeer, Roberts, Lee, Narang, Matena,
  Zhou, Li, and Liu}]{T5}
Colin Raffel, Noam Shazeer, Adam Roberts, Katherine Lee, Sharan Narang, Michael
  Matena, Yanqi Zhou, Wei Li, and Peter~J. Liu. 2020.
\newblock \href {http://jmlr.org/papers/v21/20-074.html} {Exploring the limits
  of transfer learning with a unified text-to-text transformer}.
\newblock \emph{J. Mach. Learn. Res.}, 21:140:1--140:67.

\bibitem[{Roller et~al.(2021)Roller, Dinan, Goyal, Ju, Williamson, Liu, Xu,
  Ott, Smith, Boureau, and Weston}]{roller-etal-2021-recipes}
Stephen Roller, Emily Dinan, Naman Goyal, Da~Ju, Mary Williamson, Yinhan Liu,
  Jing Xu, Myle Ott, Eric~Michael Smith, Y-Lan Boureau, and Jason Weston. 2021.
\newblock \href {https://doi.org/10.18653/v1/2021.eacl-main.24} {Recipes for
  building an open-domain chatbot}.
\newblock In \emph{Proceedings of the 16th Conference of the European Chapter
  of the Association for Computational Linguistics: Main Volume}, pages
  300--325, Online. Association for Computational Linguistics.

\bibitem[{Schulman et~al.(2017)Schulman, Wolski, Dhariwal, Radford, and
  Klimov}]{PPO}
John Schulman, Filip Wolski, Prafulla Dhariwal, Alec Radford, and Oleg Klimov.
  2017.
\newblock \href {http://arxiv.org/abs/1707.06347} {Proximal policy optimization
  algorithms}.
\newblock \emph{CoRR}, abs/1707.06347.

\bibitem[{Shang et~al.(2015)Shang, Lu, and Li}]{shang-etal-2015-neural}
Lifeng Shang, Zhengdong Lu, and Hang Li. 2015.
\newblock \href {https://doi.org/10.3115/v1/P15-1152} {Neural responding
  machine for short-text conversation}.
\newblock In \emph{Proceedings of the 53rd Annual Meeting of the Association
  for Computational Linguistics and the 7th International Joint Conference on
  Natural Language Processing (Volume 1: Long Papers)}, pages 1577--1586,
  Beijing, China. Association for Computational Linguistics.

\bibitem[{Stiennon et~al.(2020)Stiennon, Ouyang, Wu, Ziegler, Lowe, Voss,
  Radford, Amodei, and Christiano}]{summary}
Nisan Stiennon, Long Ouyang, Jeff Wu, Daniel~M. Ziegler, Ryan Lowe, Chelsea
  Voss, Alec Radford, Dario Amodei, and Paul~F. Christiano. 2020.
\newblock \href {http://arxiv.org/abs/2009.01325} {Learning to summarize from
  human feedback}.
\newblock \emph{CoRR}, abs/2009.01325.

\bibitem[{Sun et~al.(2022{\natexlab{a}})Sun, Wang, Song, Feng, and
  Nie}]{CommnetGenerationTOMM2022}
Teng Sun, Chun Wang, Xuemeng Song, Fuli Feng, and Liqiang Nie.
  2022{\natexlab{a}}.
\newblock \href {https://doi.org/10.1145/3475872} {Response generation by
  jointly modeling personalized linguistic styles and emotions}.
\newblock \emph{{ACM} Trans. Multim. Comput. Commun. Appl.}, 18(2):52:1--52:20.

\bibitem[{Sun et~al.(2022{\natexlab{b}})Sun, Wang, Song, Feng, and
  Nie}]{autoGenTOMM}
Teng Sun, Chun Wang, Xuemeng Song, Fuli Feng, and Liqiang Nie.
  2022{\natexlab{b}}.
\newblock \href {https://doi.org/10.1145/3475872} {Response generation by
  jointly modeling personalized linguistic styles and emotions}.
\newblock \emph{{ACM} Trans. Multim. Comput. Commun. Appl.}, 18(2):52:1--52:20.

\bibitem[{Touvron et~al.(2023)Touvron, Lavril, Izacard, Martinet, Lachaux,
  Lacroix, Rozière, Goyal, Hambro, Azhar, Rodriguez, Joulin, Grave, and
  Lample}]{touvron2023llama}
Hugo Touvron, Thibaut Lavril, Gautier Izacard, Xavier Martinet, Marie-Anne
  Lachaux, Timothée Lacroix, Baptiste Rozière, Naman Goyal, Eric Hambro,
  Faisal Azhar, Aurelien Rodriguez, Armand Joulin, Edouard Grave, and Guillaume
  Lample. 2023.
\newblock \href {http://arxiv.org/abs/2302.13971} {Llama: Open and efficient
  foundation language models}.

\bibitem[{Wang et~al.(2021{\natexlab{a}})Wang, Li, and
  Zheng}]{CommnetGenerationAAAI21}
Wei Wang, Piji Li, and Hai{-}Tao Zheng. 2021{\natexlab{a}}.
\newblock \href {https://ojs.aaai.org/index.php/AAAI/article/view/17647}
  {Generating diversified comments via reader-aware topic modeling and saliency
  detection}.
\newblock In \emph{Thirty-Fifth {AAAI} Conference on Artificial Intelligence,
  {AAAI} 2021, Thirty-Third Conference on Innovative Applications of Artificial
  Intelligence, {IAAI} 2021, The Eleventh Symposium on Educational Advances in
  Artificial Intelligence, {EAAI} 2021, Virtual Event, February 2-9, 2021},
  pages 13988--13996. {AAAI} Press.

\bibitem[{Wang et~al.(2021{\natexlab{b}})Wang, Li, and
  Zheng}]{autoCommentGenAAAI}
Wei Wang, Piji Li, and Hai{-}Tao Zheng. 2021{\natexlab{b}}.
\newblock \href {https://ojs.aaai.org/index.php/AAAI/article/view/17647}
  {Generating diversified comments via reader-aware topic modeling and saliency
  detection}.
\newblock In \emph{Thirty-Fifth {AAAI} Conference on Artificial Intelligence,
  {AAAI} 2021, Thirty-Third Conference on Innovative Applications of Artificial
  Intelligence, {IAAI} 2021, The Eleventh Symposium on Educational Advances in
  Artificial Intelligence, {EAAI} 2021, Virtual Event, February 2-9, 2021},
  pages 13988--13996. {AAAI} Press.

\bibitem[{Wang et~al.(2020)Wang, Ke, Zheng, Huang, Jiang, Zhu, and
  Huang}]{wang2020chinese}
Yida Wang, Pei Ke, Yinhe Zheng, Kaili Huang, Yong Jiang, Xiaoyan Zhu, and
  Minlie Huang. 2020.
\newblock \href {https://arxiv.org/abs/2008.03946} {A large-scale chinese
  short-text conversation dataset}.
\newblock In \emph{NLPCC}.

\bibitem[{Wu et~al.(2023)Wu, Antonova, Kan, Lepert, Zeng, Song, Bohg,
  Rusinkiewicz, and Funkhouser}]{DBLP:journals/corr/abs-2305-05658}
Jimmy Wu, Rika Antonova, Adam Kan, Marion Lepert, Andy Zeng, Shuran Song,
  Jeannette Bohg, Szymon Rusinkiewicz, and Thomas~A. Funkhouser. 2023.
\newblock \href {https://doi.org/10.48550/arXiv.2305.05658} {Tidybot:
  Personalized robot assistance with large language models}.
\newblock \emph{CoRR}, abs/2305.05658.

\bibitem[{Xing et~al.(2018)Xing, Wu, Wu, Huang, and Zhou}]{responseGenAAAI2018}
Chen Xing, Yu~Wu, Wei Wu, Yalou Huang, and Ming Zhou. 2018.
\newblock \href
  {https://www.aaai.org/ocs/index.php/AAAI/AAAI18/paper/view/16510}
  {Hierarchical recurrent attention network for response generation}.
\newblock In \emph{Proceedings of the Thirty-Second {AAAI} Conference on
  Artificial Intelligence, (AAAI-18), the 30th innovative Applications of
  Artificial Intelligence (IAAI-18), and the 8th {AAAI} Symposium on
  Educational Advances in Artificial Intelligence (EAAI-18), New Orleans,
  Louisiana, USA, February 2-7, 2018}, pages 5610--5617. {AAAI} Press.

\bibitem[{Yang et~al.(2019)Yang, Xu, Wu, and Li}]{yang-etal-2019-read}
Ze~Yang, Can Xu, Wei Wu, and Zhoujun Li. 2019.
\newblock \href {https://doi.org/10.18653/v1/D19-1512} {Read, attend and
  comment: A deep architecture for automatic news comment generation}.
\newblock In \emph{Proceedings of the 2019 Conference on Empirical Methods in
  Natural Language Processing and the 9th International Joint Conference on
  Natural Language Processing (EMNLP-IJCNLP)}, pages 5077--5089, Hong Kong,
  China. Association for Computational Linguistics.

\bibitem[{Zhang et~al.(2019)Zhang, Lan, Pang, Guo, and
  Cheng}]{zhang-etal-2019-recosa}
Hainan Zhang, Yanyan Lan, Liang Pang, Jiafeng Guo, and Xueqi Cheng. 2019.
\newblock \href {https://doi.org/10.18653/v1/P19-1362} {{R}e{C}o{S}a: Detecting
  the relevant contexts with self-attention for multi-turn dialogue
  generation}.
\newblock In \emph{Proceedings of the 57th Annual Meeting of the Association
  for Computational Linguistics}, pages 3721--3730, Florence, Italy.
  Association for Computational Linguistics.

\bibitem[{Zhang et~al.(2020)Zhang, Sun, Galley, Chen, Brockett, Gao, Gao, Liu,
  and Dolan}]{zhang-etal-2020-dialogpt}
Yizhe Zhang, Siqi Sun, Michel Galley, Yen-Chun Chen, Chris Brockett, Xiang Gao,
  Jianfeng Gao, Jingjing Liu, and Bill Dolan. 2020.
\newblock \href {https://doi.org/10.18653/v1/2020.acl-demos.30} {{DIALOGPT} :
  Large-scale generative pre-training for conversational response generation}.
\newblock In \emph{Proceedings of the 58th Annual Meeting of the Association
  for Computational Linguistics: System Demonstrations}, pages 270--278,
  Online. Association for Computational Linguistics.

\bibitem[{Zheng et~al.(2018)Zheng, Wang, Chen, and Sangaiah}]{autoCommentGen1}
Hai{-}Tao Zheng, Wei Wang, Wang Chen, and Arun~Kumar Sangaiah. 2018.
\newblock \href {https://doi.org/10.1109/ACCESS.2017.2774839} {Automatic
  generation of news comments based on gated attention neural networks}.
\newblock \emph{{IEEE} Access}, 6:702--710.

\end{thebibliography}

% \bibliographystylelanguageresource{lrec-coling2024-natbib}
% \bibliographylanguageresource{languageresource}

% \appendix
% \input{texFiles/appendix}

\end{document}